\documentclass{article}

    \usepackage[preprint]{neurips_2025}

\usepackage[utf8]{inputenc} 
\usepackage[T1]{fontenc}    
\usepackage{hyperref} 
\usepackage{url}            
\usepackage{booktabs}       
\usepackage{amsfonts}       
\usepackage{nicefrac}       
\usepackage{microtype}      
\usepackage{xcolor}         
\usepackage{algorithm}
\usepackage{algpseudocode}
\usepackage{amsmath}
\usepackage{multicol}
\usepackage{geometry}
\usepackage{xcolor}
\usepackage{tcolorbox}
\usepackage{csquotes}
\usepackage{siunitx}
\usepackage{graphicx}
\usepackage{booktabs}  
\usepackage{multirow}  
\usepackage{lineno}
\usepackage{subcaption}
\usepackage{epigraph}

\definecolor{darkblue}{rgb}{0, 0, 0.5}

\title{Position: Machine Learning Conferences Should Establish a ``Refutations and Critiques'' Track}

\author{%
  Rylan Schaeffer\thanks{Correspondence to \texttt{rschaef@cs.stanford.edu}, \texttt{jkazdan@stanford.edu}, \texttt{sanmi@cs.stanford.edu}, \texttt{donoho@stanford.edu}.}\\
  Stanford
  \And
  Joshua Kazdan\\
  Stanford
  \And
  Yegor Denisov-Blanch\\
  Stanford
  \And
  Brando Miranda\\
  Stanford
  \And
  Matthias Gerstgrasser
  \And
  Susan Zhang
  \And
  Andreas Haupt\\
  Stanford
  \And
  Isha Gupta\\
  ETH Z\"urich \& Stanford
  \And
  Elyas Obbad\\
  Stanford\\
  \And
  Jesse Dodge\\
  Allen Institute for AI
  \And
  Jessica Zosa Forde\\
  Brown University\\
  \And
  Francesco Orabona\\
  KAUST\\
  \And
  Sanmi Koyejo\\
  Stanford\\
  \And
  David Donoho\\
  Stanford\\
}

\begin{document}

\maketitle

\begin{abstract}
Science progresses by iteratively advancing and correcting humanity's understanding of the world. In machine learning (ML) research, rapid advancements have led to an explosion of publications, but have also led to misleading, incorrect, flawed or perhaps even fraudulent studies being accepted and sometimes highlighted at ML conferences due to the fallibility of peer review. While such mistakes are understandable, ML conferences do not offer robust processes to help the field systematically correct when such errors are made.
This position paper argues that ML conferences should establish a dedicated ``Refutations and Critiques'' (R\&C) Track. This R\&C Track would provide a high-profile, reputable platform to support vital research that critically challenges prior research, thereby fostering a dynamic self-correcting research ecosystem.
We discuss key considerations including track design, review principles, potential pitfalls, and provide an illustrative example submission concerning a recent ICLR 2025 Oral.
We conclude that ML conferences should create official, reputable mechanisms to help ML research self-correct.
\end{abstract}

\section{Introduction}
\label{sec:introduction}




Advancement of scientific knowledge is inherently a cycle of discovery and refinement \citep{kuhn1962structure}. Within machine learning (ML), the rapid proliferation of research combined with the fallibility of peer review occasionally results in dissemination and even commendation of research that is misleading, incorrect, flawed, or potentially fraudulent. While such outcomes are understandable, their occurrence is undesirable, and a critical gap exists in the current ML conference ecosystem: there is no official mechanism for critiquing problematic publications and rectifying the scientific record.

This absence of an official corrective mechanism causes problems: inaccuracies can propagate within the scientific record, fields can putter around under the illusion of progress, or critiques can be relegated to informal channels lacking rigorous evaluation.  Minimizing such problematic publications by reforming the review process is unlikely to be a solution due to inadequate incentives for reviewers and institutional comfort with peer review as an efficient and reasonably well-trusted process.

\textbf{In light of these challenges, this position paper champions the creation of a ``Refutations and Critiques" (R\&C) Track within ML conferences to provide an official mechanism for the field to self-correct.} The objective is to provide the community with a high-profile, reputable platform dedicated to the critical re-examination of prior research published in ML conferences.
Researchers would be encouraged to submit responses and criticisms of others' research as well as of their own research; authors of critiqued papers could play a critical role in the peer review process to ensure fairness and quality of submitted refutations and promote a structured scholarly exchange.

In this position paper, we elaborate on the critical facets of this proposal, including the reasoning for the ``Refutations and Critiques" designation, the necessity of a distinct track, the suitability of ML conferences as hosting venues, and the foundational principles for its peer review rubric. We also explore potential pitfalls and advise how they can be preempted. To provide a concrete understanding of envisioned submissions, we conclude by presenting an illustrative manuscript examining a recent ICLR 2025 Oral. While we fully stand behind the analysis of the example manuscript, we specifically ask readers to evaluate it as an example of what submissions to the R\&C could look like. Our hope is that this proposed track will supercharge ML research to rectify itself and continuously improve.

As an important disclaimer, certain observations in this work draw on community anecdotes and personal experience because peer-reviewed evidence on post-publication critiques is scarce. These valuable yet undocumented insights are precisely the type of scholarship our proposed R\&C Track seeks to legitimize and disseminate.
\section{The Case For a Corrective Mechanism at Machine Learning Conferences}
\label{sec:case_for_corrective_mechanisms}

We begin by establishing existing problems in the machine learning (ML) research status quo:
\begin{itemize}
    \item The fallibility of peer review means that ML conferences sometimes accept and even highlight misleading, incorrect, flawed, or sometimes fraudulent research, causing false information to proliferate (Sec.~\ref{sec:case_for_corrective_mechanisms:subsec:accept_flawed_research}).
    \item ML conferences currently lack official processes for rectifying such errors once they are part of the scientific record (Sec.~\ref{sec:case_for_corrective_mechanisms:subsec:lack_mechansisms}).
    \item Reforming the peer review process to minimize or prevent problematic research is likely impractical due to reviewer incentives and institutional trust in the status quo (Sec.~\ref{sec:case_for_corrective_mechanisms:subsec:challenges_to_reform}).
    \item Lack of formal recourse harms the field, imposes costs and relegates disputes to non-scientific venues without impartial adjudication (Sec.~\ref{sec:case_for_corrective_mechanisms:subsec:undesirable_outcomes}).
\end{itemize}

\subsection{Machine Learning Conferences Accept and Highlight Flawed Research}
\label{sec:case_for_corrective_mechanisms:subsec:accept_flawed_research}

While the peer review process is central to research, nearly every machine learning (ML) researcher has stories of personal challenges they've encountered with peer review; we compile some common complaints in Appendix~\ref{app:sec:fallibility_of_peer_review}.
However, we are most concerned with a specific complaint: failures in the review process lead to research that is misleading, incorrect, flawed or perhaps even fraudulent.
These papers are of special concern because when they receive acceptance and at times prestigious recognition, they impose significant costs on research. Specifically, such papers:
\begin{itemize}
    \item incentivize others to produce less rigorous and overly hyped research,
    \item divert attention and resources from more rigorous work,
    \item waste future researchers' time and energy building on flawed foundations,
    \item pollute the scientific record and misdirect future work,
    \item generally undermine public trust in ML research.
\end{itemize}

Over the years, there have been multiple instances of top ML papers being publicly contested by reputable researchers:
\begin{itemize}
    \item In privacy, the ICML 2022 Outstanding Paper ``Privacy for Free: How does Dataset Condensation Help Privacy?'' claimed that a technique called data condensation could be used to train performant and private models \citep{dong2022privacyfreedoesdataset}.  \citet{carlini2022freelunchprivacyfree} responded that statistical tests show the method does not improve the privacy over a naive baseline and the technique is dominated by a standard and widely used baseline \citep{abadi2016deep}.  Moreover, \citet{carlini2022freelunchprivacyfree} showed that the theoretical results are meaningless because one of the proof assumptions is stronger than the result \citep{vitalyFM2022privacy}.
    \item In post-training of language models, a foundational OpenAI publication ``Learning to Summarize from Human Feedback'' introduced an analytically-derived equality for the KL Divergence between a probability distribution and its corresponding ``Best-of-$n$'' distribution \citep{stiennon2020learning}, which allowed for studying many interesting effects such as reward model overoptimization \citep{hilton2022goodhartslaw,gao2023scaling,coste2024reward} and others \citep{bai2022traininghelpfulharmlessassistant}. However, the equation for the KL divergence was actually a loose upper bound \citep{beirami2025theoreticalguaranteesbestofnalignment,mroueh2024information}. To the best of our knowledge, the original studies have not been re-examined to see how correct calculation of the KL Divergence changes the original results (if at all).
    \item In parameter-free optimization, the ICML 2023 Outstanding Paper ``Learning-Rate-Free Learning by D-Adaptation'' \citep{defazio2023learningratefreelearningdadaptation} exhibited a ``new method'' for parameter-free optimization of Lipschitz functions, and offered provable guarantees of convergence.
    \citet{orabona2023yetanothericmlawardfiasco} claimed that the core contribution of the paper was established eight years prior, and stronger guarantees for similar algorithms already existed in the literature and  \citet{defazio2023learningratefreelearningdadaptation}.
    Therefore, \citet{orabona2023yetanothericmlawardfiasco} claimed that, when viewed in light of prior work, \citet{defazio2023learningratefreelearningdadaptation} exhibited no added novelty.
    \item In neuroscience-inspired AI, a Nature publication \citep{banino2018vector} and two NeurIPS Spotlights \citep{sorscher2019unified,nayebi2021explaining} claimed that neural representations matching those found in the mammalian brain emerge naturally in artificial neural networks trained on a task called path integration \citep{wikipediaPathIntegration}. However, \citet{schaeffer2022nofreelunch,schaeffer2023disentanglingfactgridcell} showed that such neural representations did not emerge due to the task or other relevant constraints, but were instead inserted into the artificial networks by the researchers in subtle ways. Moreover, the way in which the neural representations were inserted contradicted key properties of the relevant biological neural circuits \citep{schaeffer2023testingassumptionsunderlyingunified}.
    \item Not all refutations grow out of mistakes or misconduct; some involve differences in scientific assumptions and experimental methodologies that vastly change projected outcomes.  In model-data feedback loops, a Nature publication \citet{shumailov2024ai} and an assortment of well-cited conference papers \citep{alemohammad2023selfconsuminggenerativemodelsmad,dohmatob2024modelcollapsedemystifiedcase} promoted the viewpoint that future deep generative models models will swiftly and drastically degrade as they are trained on data generated by earlier generative models. However, \citet{gerstgrasser2024modelcollapseinevitablebreaking,kazdan2025collapsethriveperilspromises,dey2024universality} showed that these papers unrealistically trained new models exclusively on data from the most recent generative model, which is tantamount to throwing away the internet and training GPT-5 exclusively on the outputs of GPT-4; when models are instead trained on a more realistic mixture of real and synthetic data, models demonstrate little-to-no degradation over time.  
\end{itemize}

There are, of course, many additional examples that could be mentioned, e.g., \citep{maini2025reassessingemnlp2024s, rando2025donotwrite, chandak2025llmrlincorrect, golechha2025intricacies, ivanova2025towardsmorerigorous}.
In cases like these, where a refutation is written, community members can more easily track the scientific conversation. 
In other cases, it is more difficult.
To give a prominent example, \citet{le2014distributed}'s ``Doc2Vec'' has been cited nearly \num{14,000} times, but the second author has publicly stated that the results by the other author were not reproducible \citep{stackexchangeAnswer222501}; however, finding this out requires laboriously stitching together comments scattered across the internet \citep{hackernewsDoc2Vec,stackexchangeDoc2VecQuestion} to gain a comprehensive understanding; such effort was done by \citet{raff2023siren}.

At this point, one could reasonably criticize the above points for being non-academic un-scientific hearsay from the internet.
Some of the above responses to existing papers are blog posts or Twitter threads or preprints, which limits the authority of findings due to lack of scrutiny by independent experts. At the same time, there is no clear place for such contributions in ML conferences.

\subsection{Machine Learning Conferences Lack Mechanisms for Rectifying the Scientific Record}
\label{sec:case_for_corrective_mechanisms:subsec:lack_mechansisms}

When published papers are later identified as misleading, incorrect, flawed or fraudulent, the largest ML conferences generally lack processes to address such situations.
ML conferences lack editorial boards with the prerogative to remove clearly flawed papers post-acceptance, and also lack mechanisms for authors or trusted parties to attach corrigenda or errata.
For egregiously flawed papers where retraction is appropriate, nothing can compel authors to withdraw their work, and voluntary withdrawal after acceptance is culturally abnormal in ML, even when mistakes are acknowledged.\footnote{Authors cannot be compelled to withdraw after acceptance based on the 2025 guides for reviewers, area chairs, and senior area chairs \citep{neurips2025ac,neurips2025sac,neurips2025reviewer}.}
Lastly, most ML conferences offer no official mechanism for reputable researchers to publicize discovered mistakes or problems alongside publications; the closest example is ICLR permitting public comments via OpenReview, but as we argue in Sec~\ref{sec:case_for_corrective_mechanisms:subsec:undesirable_outcomes}, public comments have low visibility, carry little weight and  again lack verification by independent experts.
The structure of ML conferences, as well as their transience, produces these problems: unlike journals that can force retraction or post errata, ML conferences leave the decision to correct flawed work up to the authors.

\subsection{Reforming Peer Review at ML Conferences to Prevent Flawed Research Is Impractical}
\label{sec:case_for_corrective_mechanisms:subsec:challenges_to_reform}

Like most processes, peer review at ML conferences is fallible, but meaningfully overhauling it to minimize or prevent publication of flawed research faces obstacles on at least three fronts:

\paragraph{Inadequate Incentives to Improve Review Quality} Peer review at ML conferences often lacks incentives since reviewers are uncompensated volunteers or conscripts fulfilling reciprocal reviewing requirements.
To the best of our knowledge, there are only two incentives to encourage high-quality reviews, which come in the form of a carrot and a stick.  The carrot is the ``Best Reviewer Award", which recognizes outstanding reviewers with free conference registration.  The stick is the ``Responsible Reviewing Initiative," which desk-rejects authors' papers if their reciprocal reviews are low quality \citep{samleeICMLDeskReject}. This consequence is infrequently enforced, and whether it can make a meaningful difference at scale remains to be seen \citep{koniusz2025responsiblereviewing}.

\paragraph{Inadequate Time for Rigorous Scrutiny}
The fast-paced timeline for peer review at ML conferences means that even the most meticulous and motivated reviewers would still only have several weeks to scrutinize submissions.
In our experience, discovering deep flaws and conclusively demonstrating their existence and significance properly takes much longer than several weeks.  A refutations track would allow invested researchers to scrutinize existing published work over a longer time-frame before publishing a critique in a future conference.  

\paragraph{Insufficient Number of Qualified Reviewers}
This year, NeurIPS received over 27\,000 submissions \citep{ctol2025neurips}, a number that has multiplied ten times in less than ten years.  The number of submissions has increased far faster than the population of qualified PhD students, professors, and industry researchers who can provide trustworthy reviews of these papers.  Area chairs (ACs) have noted the growing difficulty of assigning papers to qualified reviewers \citep{reddit_neurips2024}; inevitable failures to do so have led to the acceptance of more faulty research.

\paragraph{Lack of a Better System} Despite complaints, ML conference peer review is a time-tested process, and we are unaware of more reliable alternatives for vetting large volumes of papers on a contracted timeline. 
Consequently, we will argue that post-hoc mechanisms for rectifying the scientific record present a more practical solution than overhauling the imperfect-but-time-tested reviewing system (Sec.~\ref{sec:proposal}).

\subsection{Lack of Mechanisms for Refuting or Correcting Flawed Publications Hinders the Field}
\label{sec:case_for_corrective_mechanisms:subsec:undesirable_outcomes}

Without appropriate channels to challenge published research, researchers are left with several suboptimal options. We enumerate each option below with corresponding anecdotes, but first note that there are likely two sides to each anecdote, but without undergoing scrutiny by independent experts, where the truth lies may be unclear.

\paragraph{Silence $\rightarrow$ Persistence and Propagation of Errors} Without an avenue to express valid criticisms, they go unvoiced or unpublished, thereby propagating incorrect knowledge that can misdirect subsequent research. 
For one recent example, \citet{liu2024sophiascalablestochasticsecondorder} proposed a novel optimizer claiming to improve over a widely-used baseline optimizer \citep{kingma2014adam,loshchilov2019adamw}, but the baseline was improperly tuned \citep{keller2024improvingadamw}. The researcher who discovered this error publicly said that clarifying this detail via a publication would not be valued by the research community and consequently never documented the finding outside of Twitter \citep{keller2025evidence}.

\paragraph{Speak Only When Harms Can No Longer Be Ignored $\rightarrow$ Damage Is Already Done} 
If the consequences of problematic research become too overwhelming to ignore, then some may feel compelled to speak.
One prominent example is \citet{agarwal2021deep}, which questioned whether the field of deep reinforcement learning (RL) was actually making progress.
The authors evaluated over 16 high profile publications from up to 6 years prior and concluded that many deep RL algorithms were not improvements as had been previously claimed.
While the contribution was extremely valuable, the field had lost significant time and resources based on a false illusion of progress.

\paragraph{Formal Public Comments on OpenReview $\rightarrow$ Low Visibility and Little Weight}
ICLR permits researchers to comment publicly during the review process.
While valuable, such public comments carry insufficient weight to formally amend the scientific record and possess inadequate visibility to alter the scientific record.
For a recent example, \citet{schaeffer2024strongmodelcollapse} demonstrated that an ICLR 2025 submission intentionally suppressed contradictory scientific evidence; the Area Chair then rejected the submission, but was overruled, and the submission was selected for a Spotlight.
For another example regarding the NeurIPS 2024 Best Paper Runner-Up, see \citet{kirsch2024rholossattribution}.

\textbf{Non-Traditional Publication Venues $\rightarrow $ Insufficient Visibility} Researchers can submit articles to less traditional publication venues such as Distill \citep{distill2016} or to newer, relatively-infrequent and less well-known venues such as the Machine Learning Reproducibility Conference (MLRC). Though these are respected and peer-reviewed venues, they are younger and do not yet have the same visibility as the large ML conferences.
As an example, Distill published a critique of the paper ``Adversarial Examples are not Bugs, They are Features''~\citep{ilyas2019adversarialexamplesbugsfeatures} called ``Adversarial Examples are Just Bugs, Too''~\citep{nakkiran2019adiscussion}, which has just 2\% the number of citations as the original paper.

\paragraph{Informal Public Comments on Blogs/Social Media $\rightarrow$ Digital Frontier Justice}
When robust mechanisms for scientific correction are absent, researchers may resort to informal channels like blogs or social media to voice refutations, including some examples shared in Sec.~\ref{sec:case_for_corrective_mechanisms:subsec:accept_flawed_research} as well as in \citet{kirsch2022bayesianmodelselection}.
However, this ``digital frontier justice'' is fraught with problems that undermine genuine scientific discourse.
Critiques aired on these platforms bypass scrutiny by independent experts and lack archival permanence crucial for the scientific record.
Instead of fostering reasoned debate, these forums (especially social media) can become popularity contents, where debates are swayed by an individual author's personal reach rather than the scientific validity of their argument. Scientific disagreements often devolve into ad hominem attacks, creating uncertainty as to whether the criticisms are based on legitimate academic concerns or personal grievances.
Additionally, these informal debates exclude researchers who are not active on these specific platforms, narrowing the scope of discussion and potentially reinforcing biases.

Such situations create a damaging double-bind for the scientific community: On one hand, unvetted accusations can unfairly tarnish the reputations of researchers who have produced valid and original scientific work. On the other, authors who publish flawed research can exploit an inherent asymmetry: their incorrect work carries the imprimatur of peer review, while the informal critiques against it do not. This allows them to simply deny or dismiss valid concerns until public attention wanes.

The consequences of this uncertainty are significant. It delays scientific progress as the community struggles to discern truth from noise. This burden falls disproportionately on early-career researchers. Lacking the extensive experience to easily differentiate robust findings from flawed ones, they may waste valuable time and resources trying to build upon an unsound foundation. Speaking from experience, the inability to reproduce work originating from renowned labs or championed by prominent scientists can lead to unfair criticism and profound self-doubt among junior researchers.

\section{Proposal: Establish a ``Refutations and Critiques'' (R\&C) Track}
\label{sec:proposal}

Rather than seeking to change peer review at ML conferences, we instead aim to leverage peer review as a familiar and reasonably trusted process to equip the field of ML with an official, reputable, and equally prestigious mechanism to address, critique and/or rectify flawed prior research.

\subsection{Overview of Proposed ``Refutations and Critiques'' (R\&C) Track}

 We propose the establishment of a dedicated ``Refutations and Critiques'' (R\&C) track within ML conferences such as NeurIPS, ICML, and ICLR. 
 The objective of this R\&C Track is to provide a high-profile, reputable, and rigorously peer-reviewed platform for research that identifies, analyzes, and corrects misleading, incorrect, or potentially fraudulent claims presented in impactful ML publications.
Officially integrating critical scholarship into main conferences would recognize this work as an indispensable component of the scientific process and would serve to enhance the overall integrity and reliability of contributions to the field of machine learning.
 Our vision for the R\&C Track is to cultivate a dynamic and robustly self-correcting research ecosystem for the ML research community.
 This track would naturally be integrated with existing conference structures, including timelines and review processes, and would likely include a pilot to evaluate whether the proposed track is effective.

\subsection{Why the Name ``Refutations and Critiques?''}
\label{sec:proposal:subsec:why_rnc}


The choice of the name ``Refutations and Critiques'' over alternatives such as ``Reproducibility'' reflects a broader scope and a stronger stance on the nature of critical scientific discourse.
While issues of reproducibility are undeniably important and would fall under the purview of this track, the title ``Refutations and Critiques'' is intended to encourage a more diverse level of engagement with prior work.
Situations arise where the original results may be reproducible and technically correct, but the surrounding narrative may be (even unintentionally) incomplete, misleading or incorrect.
Such situations have appeared across multiple topics, ranging from machine learning security \citep{carlini2017magnetefficientdefensesadversarial,athalye2018obfuscatedgradientsfalsesense,carlini2021privatelearningpossibleinstance,carlini2022freelunchprivacyfree,tramèr2024positionconsiderationsdifferentiallyprivate} to deep reinforcement learning \citep{agarwal2021deep} to model collapse \citep{gerstgrasser2024modelcollapseinevitablebreaking,kazdan2025collapsethriveperilspromises,dey2024universality,schaeffer2025positionmodelcollapsedoes} to predictable scaling of language models \citep{schaeffer2025mirage, schaeffer2025largelanguagemonkeyspower} to biological neural representations for spatial navigation \citep{schaeffer2022nofreelunch,schaeffer2023sslgc}, to meta-learning \citep{tian2020rethinkingfewshotimageclassification, raghu2020rapidlearningfeaturereuse, miranda2022curselowtaskdiversity, miranda2023pretrainingtrulybettermetalearning, chen2020closerlookfewshotclassification}.
\nocite{schaeffer2023testingassumptionsunderlyingunified}
\nocite{schaeffer2023disentanglingfactgridcell}
\nocite{spencer2022deconstructing}
\nocite{beirami2025theoreticalguaranteesbestofnalignment}
\nocite{schaeffer2025predictingdownstreamcapabilitiesfrontier}
\nocite{kirsch2022bayesianmodelselection}
To use model collapse as a recent example, many previous papers made a single critical assumption, and based on that assumption, drew strong conclusions that future generative models are doomed.
While the conclusion indeed follows from the stated assumption, the assumption itself is arguably highly unrealistic, and subsequent critical work demonstrated that by adjusting the assumption to be more realistic, the predicted collapse of future generative models is substantially mitigated or disappears entirely \citep{gerstgrasser2024modelcollapseinevitablebreaking,kazdan2025collapsethriveperilspromises,dey2024universality}.
Such research moves beyond a simple reproducibility check to a substantive critique of the generality of prior works' conclusions.

A related reason why ``refutations and critiques'' are appropriate is that if a paper lacks definitions or concrete claims, then refutations or failed reproductions have no meaning.
This point was made by \citet{carlini2020instahide}, who challenged an ML privacy paper called InstaHide and wrote:
``One of the core tenets of modern science [...] is that claims should be refutable [...] Unfortunately, InstaHide does not make falsifiable claims. [...] It defines an algorithm, and says it is private, without ever saying what that means.
As a result, it's impossible to ever write a paper that claims to break it, because defining an attack necessarily requires a definition to break.''

One may wonder: If prior critical work has been published, why then is a standalone track merited? We return to answer this question below in Section~\ref{sec:proposal:subsec:why_standalone_track}.

Thus, while reproducibility assessments are welcome as submissions, this R\&C Track aims to construct a broader aegis to explicitly encourage submissions that seek to rectify the field.
We seek to foster a space for rigorous, evidence-based challenges to prior research, whether they pertain to inadequate reproducibility, problematic assumptions, flawed experimental methodologies, incorrect analyses, misleading interpretations, or erroneous conclusions.

\subsection{Why Is A Standalone Track Merited?}
\label{sec:proposal:subsec:why_standalone_track}

In the past, critical research has been submitted to and accepted at ML conferences' Main Tracks.
Why is a dedicated R\&C Track merited?
There are several reasons:

\paragraph{Reduce Reviewer Indifference and Opposition}

Although critical responses can be accepted, e.g., \citet{santurkar2018does} and even awarded, e.g., \citet{agarwal2021deep}, such contributions are oftentimes harder to get through the peer review process.
In many cases, reviewers or area chairs are indifferent to the prior work being criticized, and so are unlikely to find the results compelling and even less likely to find the results worthy of recognition.
As one anecdote, \citet{kirsch2024does} submitted to TMLR (instead of ML conferences) because TMLR explicitly invites reproducibility studies, and even then had to push back forcefully against reviewer indifference to be accepted.
In less common cases, reviewers or area chairs are affiliated with the prior work being criticized, meaning the reviewers are no longer neutral evaluators and submitting authors can face strong opposition.

\paragraph{Create Bespoke Reviewing Standards} Responses and critiques have unique aspects that are best served via unique standards of evaluation. For instance, additional specific checks may be required (e.g., whether the criticized paper's authors are aware of the objections and have been given proper opportunity to respond) and double-blind review may not be realistic (e.g., if public interactions such as GitHub issues are referenced for evidence).

\paragraph{Link Accepted Papers to Prior Work}
A response or a critique is more than a citation; it should be explicitly linked to the publication being critiqued, similar to how corrigenda and errata are.

For a guiding example, we can look to the NeurIPS Datasets and Benchmarks (D\&B) Track~\citep{vanschoren2021neuripsdnb}.
Prior to the creation of the D\&B track, datasets and benchmarks were submitted to the NeurIPS Main Track; however, NeurIPS acted to carve out a unique track for reasons that echo what we have stated here.
NeurIPS recognized that datasets and benchmarks play a fundamental role in ML research, and that inadequate and insufficiently rigorous incorporation of datasets and benchmarks into the research process was incurring harms.
NeurIPS also clearly stated, ``there are currently not enough incentives at NeurIPS to work and publish on data and benchmarks, as evidenced by the lack of papers on this topic,'' and that reviewing criteria for the Main Track may be inapplicable for the unique considerations applicable to datasets \& benchmarks.
In our opinion, our thesis for the creation of the R\&C Track closely echoes \citet{vanschoren2021neuripsdnb}'s explanation of the decision process that led to the creation of the D\&B Track.

Therefore, the establishment of a dedicated R\&C Track is not merely about rectifying past mistakes; it is a forward-looking initiative designed to foster a culture of accountability and correction that proactively incentivizes more rigorous and higher-quality research from the outset.

\subsection{Why ML Conferences Should Host This Track?}

We advocate for ML conferences like NeurIPS to pioneer this R\&C Track for several reasons:

\paragraph{Conferring Legitimacy}

The considerable prestige of NeurIPS can legitimize R\&C research as an essential scientific contribution.  Hosting such a track at NeurIPS would offer a respected platform for this vital work, fostering an accredited self-correcting research ecosystem. 

\paragraph{Linking Accepted Papers to Criticized Papers} NeurIPS can leverage its unique position to push to link R\&C papers directly with the original publications they address.
This integration would help ensure that critiques are accessible alongside the original research, highlighting their unique role more akin to errata than subsequent research.

\paragraph{Leadership to Set Precedent} As a premier ML conference, NeurIPS is uniquely positioned to influence other major conferences like ICML and ICLR to adopt similar critical evaluation frameworks, thereby enhancing the overall scientific process. 

\paragraph{Bridge the Gap with Journals} ML conferences are preferred by far over ML journals as publication venues. This is due to a number of advantages, such as faster review and publication, better visibility, and larger impact.
However, journals have standardized ways to submit critical comments on papers published by the journal; see, for example, \citet{loosli2007comments}.
Creating an R\&C Track at ML conferences would bridge this gap.



\subsection{What Principles Should Guide the Evaluation of R\&C Submissions?}
\label{sec:evaluation}

The success of the R\&C track will necessitate tailored review criteria, distinct from those for regular research papers. These criteria will specifically guide reviewers to evaluate submissions based on the rigor of their critical analysis, the substance of the issues raised, and the constructive nature of the contribution, ensuring a focus on strengthening the scientific process.

\paragraph{Correct, Rigorous and Meticulous} Submissions must demonstrate methodological soundness. Critiques involving experiments should ensure reproducibility and transparent analysis; theoretical arguments must be logically coherent. All claims require strong, verifiable evidence (e.g., experimental results, proofs, data re-analysis) allowing for independent assessment by the reviewers.

\paragraph{Substantive} Submissions must address substantive aspects of the original work, such as its central claims, core methodologies, or foundational assumptions. The critique should have meaningful implications, clearly articulating why the identified issues (e.g., unsupported conclusions, flawed assumptions or results, limited applicability of results) are important to the broader research community. Submissions that solely challenge attribution errors or nitpick minor details should be rejected.

\paragraph{Constructive} Submissions must aim to correct and improve the scientific record and positively guide future research, rather than being solely dismissive or punitive or quarrelsome. Submissions should articulate their contribution to collective understanding, potentially offering corrections, alternative interpretations, or suggestions for better practices, all while maintaining a professional tone. Submissions must focus on the research and not on the people who conducted it.

\paragraph{Significant} Submissions should be judged based on how influential the criticized work is and how significantly the submissions change the field's understanding. Responses or critiques that substantially shift widespread, deeply held important beliefs represent the most consequential forms of scientific correction, and the R\&C track should recognize and reward such contributions.

\section{Alternative Viewpoints}

While we advocate strongly for the establishment of an R\&C Track, it is crucial to acknowledge and address alternative perspectives regarding its utility and implementation.

\paragraph{Existing Mechanisms Are Already Sufficient for Addressing Flawed Research}

As mentioned earlier, truly significant refutations or corrections can get published in the Main Tracks of ML conferences, e.g., \citet{agarwal2021deep}. From this perspective, a dedicated track might be seen as an unnecessary formalization of what already naturally happens. However, as previously discussed, these existing mechanisms have notable shortcomings and fail to recognize that any scientific discipline requires a lively and moderated discussion on the issues and errors present in previous papers.

\paragraph{R\&C Track Encourages Frivolous, Adversarial or Otherwise Unconstructive Submissions}

Another reasonable concern is the potential for the R\&C Track to encourage frivolous submissions or foster a more adversarial research culture. Critics might fear that such a track could become a venue for minor nit-picking, ad hominem attacks disguised as scientific critique, or an increase in hostile interactions between research groups. These concerns could be amplified if authors of highly-cited papers become targets of coordinated criticism campaigns or if the review process fails to maintain appropriate standards.

These are serious considerations that must be central to the design and governance of an R\&C Track. The guiding principles outlined in Section \ref{sec:evaluation} - demanding that submissions be ``Rigorous and Meticulous,'' ``Substantive,'' ``Significant'' and ``Constructive'' - are specifically intended to mitigate these risks. Additionally, the review process should include mechanisms for authors of critiqued papers to provide responses, similar to journal comment/response systems.

\paragraph{R\&C Track Papers Might Themselves Be Misleading, Incorrect, Flawed or Fraudulent}

There is also the risk that refutations themselves could be flawed, leading to the unjust discrediting of sound research.  For example, the refutation \citep{zhang2025evaluatingrobustnessensembleeverywhere} of ``Ensemble Everything Everywhere" initially claimed that \citet{fort2024ensembleeverywheremultiscaleaggregation} was not robust.  The authors of the refutation later discovered errors in their own experiments that caused them to revise their claims.
To address this concern, the R\&C Track would need rigorous review standards and potentially multiple rounds of author-reviewer interaction to ensure accuracy.
Authors of the original work should likely be invited to participate in a unique capacity to ensure fairness of the process.
\section{Refutations \& Critiques Track Is Complementary to Related Efforts}

Our proposed R\&C Track builds upon and complements existing reproducibility efforts.
One of the earliest and most prominent systematic efforts to encourage reproducibility in ML was \citet{pineau2017iclr}'s ICLR 2018 Reproducibility Challenge, which was repeated in ICLR 2019 \citep{pineau2019iclrreproducibilitychallenge} and has led to poster sessions and workshops as well as the Machine Learning Reproducibility Challenge (MLRC)'s in-person conference\footnote{\url{https://reproml.org/blog/announcing_mlrc2025/}}.  Machine Learning Retrospectives was hosted in 2020 at ICML and NeurIPS, which invited authors to expound upon their own previous papers, correcting mistakes and inaccuracies \citep{ml_retrospectives}.  It also accepted submissions that highlighted poor scientific practice in a particular area, pointed out conflicting claims in related papers, or documented changes in the consensus over time.
Transactions on Machine Learning Research (TMLR) additionally invites reproducibility submissions that it awards with a Reproducibility certification.
However, these efforts focus primarily on reproducibility verification, while our proposed R\&C Track has a broader scope, such as accepting papers that refute misleading conclusions drawn from reproducible experiments, challenge assumptions, or correct mis-interpretations.
By integrating with major ML conferences, the R\&C Track would bring additional prestige, visibility, and submission opportunities to these important efforts.

Other fields, such as biological and social sciences, have
recognized struggles with reproducibility and established mechanisms to combat faulty science.  For instance, the Journal of Comments and Replications in Economics (JCRE) exists to examine the reproducibility of past Economics work and to explores whether published results are correct, robust, and generalizable. 
Noting that many highly cited articles in psychology and psychiatry do not publish accompanying data, \citet{hardwicke2018populating} founded the Data Ark initiative to archive data from seminal studies.
Moreover, many papers examine aggregate statistics on reproducibility and policies to address the reproducibility crisis~\citep{klein2018manylabs2, hardwicke2022post, 10.7554/eLife.72185, 10.7554/eLife.71601}.  

\section{An Illustrative Example Submission to Our Proposed R\&C Track}
\label{sec:illustrative_example}

To demonstrate what submissions to our proposed R\&C Track might look like, we concurrently release a standalone paper \citep{schaeffer2025minpmaxexaggerationcritical} that critically analyzes a recent ICLR 2025 Oral \citep{nguyen2024minp} and concludes, based on evidence presented by the paper and additional experiments conducted using the paper's code, that the evidence fails to support the paper's central claim.
While we fully stand behind the analysis of the example manuscript, \textbf{we specifically ask readers to evaluate it as an example of what submissions to the R\&C Track might look like.}

\section{Conclusion}
\label{sec:conclusion}

The pressure in ML to quickly publish research that achieves new benchmark records can lead some researchers to take shortcuts or make errors while racing to meet publication deadlines.  The conference peer review process, which occurs on a contracted timeline relative to academic journals, inevitably admits flawed work to top venues.  Due to the difficulty of reforming the peer review process, we argue that an R\&C track would incentivize more honest, higher quality work.  Currently, discussions around academic integrity and reproducibility take place on unmoderated forums like Twitter, allowing scientific discussions to devolve into personal attacks.  Consequently, the public as well as other scientists remain unsure of the truth, and new work tries to build upon faulty prior conclusions.  Peer-reviewed, high-quality refutations currently receive insufficient attention, making them easy to miss when researching a topic.  Large ML conferences have inadvertently created the conditions that breed flimsy experimentation and corner-cutting: we offer a mechanism by which mistakes can be caught and rectified.  An R\&C track would add professionalism and visibility to the process of refuting prior work.  It would streamline the scientific process, reducing wasted time and resources expended reproducing faulty work.  Finally, it would incentivize scientists to produce higher-quality papers, and pave the way to a more open scientific discussion.

In a field like machine learning, where the technologies we build are quickly integrated into day-to-day life, producing a clean scientific record is of paramount importance.  Mistakes can result in massive waste of finite resources, propagate false impressions of AI safety, and incur broad societal risks.  To limit the growing potential for harm, conferences should act now.
\clearpage

\bibliography{references_rylan}
\bibliographystyle{colm2025_conference}


\appendix

\clearpage
\section{Additional Evidence of the Fallibility of the Machine Learning Conference Peer Review Process}
\label{app:sec:fallibility_of_peer_review}

Conferences sometimes (nearly) reject seminal papers: NeurIPS 2014 rejected Knowledge Distillation \citep{hinton2015distilling} in 2014 \citep{vinyals2019knowledge}, Adam \citep{kingma2014adam} was originally rejected from ICLR 2015 before the decision was overturned, and distributed Shampoo~\citep{anil2020scalable} was rejected at ICLR 2021
before later winning the AlgoPerf optimization track \citep{arohan2024shampoo}.
Recently, the senior author of ICLR 2025's Best Paper \citep{qi2024safetyalignmentjusttokens} shared publicly that the paper was rejected from NeurIPS 2024 even though their resubmission had not changed ``the key contributions in any significant manner" \citep{mittal2024resubmission}.
Other stories include frustrating interactions with obstinate reviewers, or on the flip side, aggressive authors ``wearing down" reviewers \citep{icmlpccharis2025reviewing}.
Two well-known experiments at NeurIPS, one in 2014 \citep{cortes2021inconsistencyconferencepeerreview} and another in 2021 \citep{beygelzimer2021neurips2021}, have sought to quantify the noise in the ML conference review process.
In short, the review process is, at times, fallible.




\end{document}